\documentclass[conference]{IEEEtran}
\IEEEoverridecommandlockouts
\usepackage{cite}
\usepackage{amsmath,amssymb,amsfonts}
\usepackage{graphicx}
\usepackage{textcomp}
\usepackage{xcolor}
\def\BibTeX{{\rm B\kern-.05em{\sc i\kern-.025em b}\kern-.08em
    T\kern-.1667em\lower.7ex\hbox{E}\kern-.125emX}}

\usepackage{subcaption}
\usepackage{color,colortbl}
\usepackage{xspace}
\usepackage{hyperref}
\hypersetup{
  colorlinks   = true, %Colours links instead of ugly boxes
  urlcolor     = blue, %Colour for external hyperlinks
  linkcolor    = blue, %Colour of internal links
  citecolor   = blue, %Colour of citations
  breaklinks = true %use \href{url}{\nolinkurl{display URL}} for breakable long URLs
}
\usepackage{algorithm,algorithmicx}
\usepackage{hhline}

\definecolor{Gray}{gray}{0.9}

\newcommand{\ie}{\textit{i.e.,}\xspace}
\newcommand{\bw}{StarCraft~1\xspace}
\newcommand{\lotv}{StarCraft~2\xspace}
\newcommand{\taunt}{\textsc{Taunt}\xspace}

\newcommand{\cspmodel}[3]% 
{\begin{trivlist}
  \item[]%
    \textbf{Variables:} #1\\
    \textbf{Domains:} #2\\
    \textbf{Constraints:} #3
  \end{trivlist}%
}

\newcommand{\copmodel}[4]% 
{\begin{trivlist}
  \item[]%
    \textbf{Variables:} #1\\
    \textbf{Domains:} #2\\
    \textbf{Constraints:} #3\\
    \textbf{Objective function:} #4
  \end{trivlist}%
}

\newcommand{\objective}[1]% 
{\begin{trivlist}
  \item[]%
    \textbf{Objective functions:} #1
  \end{trivlist}%
}

\begin{document}

\title{Terrain Analysis in StarCraft~1 and~2 as Combinatorial Optimization}

\author{\IEEEauthorblockN{Florian Richoux}
\IEEEauthorblockA{\textit{AIST} \\
Tokyo, Japan\\
florian@richoux.fr}
}

\maketitle

\begin{abstract}
  Terrain analysis in Real-Time Strategy  games is a necessary step to
  allow spacial reasoning.  The goal  of terrain analysis is to gather
  and process  data about the  map topology  and properties to  have a
  qualitative spatial representation. On StarCraft games, all previous
  works  on  terrain  analysis  propose  a  crisp  analysis  based  on
  connected  component  detection,  Voronoi  diagram  computation  and
  pruning, and region merging. Those  methods have been implemented as
  game-specific libraries,  and they can  only offer the same  kind of
  analysis for  all maps and all  users.  In this paper,  we propose a
  way  to consider  terrain analysis  as a  combinatorial optimization
  problem.  Our method allows different  kinds of analysis by changing
  constraints or the objective function  in the problem model. We also
  present  a library,  \taunt,  implementing our  method  and able  to
  handle both \bw  and \lotv maps. This makes our  library a universal
  tool for StarCraft bots with different spatial representation needs.
  We believe our library unlocks the possibility to have real adaptive
  AIs playing StarCraft,  and can be the starting point  of a new wave
  of bots.
\end{abstract}


\section{Introduction}

From the  top-left corner of the  map, 2 minutes 38  seconds after the
beginning of  the game, the  blue player sends two  \emph{stalkers}, a
basic  fast-moving  unit  of  the Protoss  faction,  towards  the  red
player's base.  Red has the same number of stalkers to defend himself,
but  Blue's  micromanagement  is  so  efficient  that  the  battle  is
completely at Blue's advantage.

Red is defending  well, and he managed to  produce \emph{immortals}, a
strong unit  known to be  a hard counter  against an army  composed of
stalkers. However,  Blue is continuously, tirelessly  sending waves of
stalkers  against Red's  army,  and combined  with  an excellent  unit
control, has overwhelmed the opponent that has no other choice than to
resign.

Despite the headlong, stubborn strategy of Blue, who did not scout the
opponent and was  then rolling this strategy out  blindly, and despite
the  good  strategical  decisions  from  Red,  Blue's  domination  was
indisputable. This was the third  game between the professional player
MaNa, in  red, versus  Alphastar, in  blue, on  December 19,  2018. In
fact, this  describes most of the  games played by Alphastar  that day
and  some months  later  on  the \lotv  ladder:  Alphastar starts  and
finishes a game  with the same strategy rolled out  at the perfection,
without any significant adaptations regarding its opponent's strategy.

But Alphastar is not an exception  in StarCraft AI: the large majority
of    \bw   and    \lotv    bots   have    scripted   or    unadaptive
behaviors\cite{Certicky2019}.    Despite    being   potentially   very
effective, those bots  are somehow getting around  the difficulty, and
thus the interest,  of RTS games used as AI  benchmarks, by exploiting
those benchmarks' flaws and imperfections.

Several directions  are possible for  making StarCraft bots  with some
real capacities of adaptability.  One  way is to design a non-scripted
bot, able  to play  the three  races on  \bw and  \lotv with  the same
high-level decision-making mechanisms.

However, to  enable homogeneous  spatial reasoning  over both  \bw and
\lotv, bots need a terrain analysis tool working the same way on these
two games.   Ideally, such  a tool  should be  flexible enough  to fit
different needs in  terms of terrain analysis, to be  exploited at the
best by different bots applying different methods.

This is the starting  point of this work. In this  paper, we propose a
flexible method to tackle terrain  analysis in both StarCraft games as
a  combinatorial  optimization  problem,  that  can  be  expressed  by
different models  supporting different analyses. After  presenting our
method  and our  combinatorial  optimization problem  models, we  show
experimental  results of  the C++  library we  developed, tested  upon
diverse \bw and \lotv maps.

\section{Background}

\subsection{RTS games and StarCraft}

A Real-Time Strategy game, or RTS game, is a genre of video game where
players must gather resources to develop an army, and must destroy the
army  or  the buildings  of  other  players  to win.   Unlike  classic
strategic  board games  such as  Chess, players  are not  making moves
alternatively, by simultaneously in real-time. The most popular way to
play this kind of games involves two players in a 1 vs 1 game.

This  study tackles  terrain analysis  on the  two widely  popular RTS
games of  the StarCraft series:  the first one  ``StarCraft'' together
with its extension ``StarCraft: Brood  War'', denoted in this paper as
\emph{StarCraft 1}, and the second one ``StarCraft II'' under its last
version named ``Legacy of the  Void'', denoted here as \emph{StarCraft
  2}. To explain something concerning both games, we will simply write
\emph{StarCraft}.   Both  games  have  a similar  gameplay  with  some
different  features,  such  as   unit  characteristics  and  some  map
properties.

In StarCraft, players must gather two types of resources (minerals and
gas) in order to construct buildings.  Those buildings are required to
produce units, to improve some unit characteristics or even to add new
ones.  All of these have a given cost to be paid in resources.

Games are played  on a rectangular map, discretized  into tiles called
\emph{buildable} tiles,  upon which buildings can  be constructed, and
smaller tiles called \emph{walkable} tiles,  which are discrete in \bw
but continuous values  in \lotv (so not tiles per  se), on which units
can walk. Indeed,  some pixels on a map can  be walkable without being
buildable.  This  mostly happens at  the edge of some  natural borders
such as cliffs or water.

Like most RTS games, the map is covered by a ``fog of war'' preventing
players  seeing   enemy  units  that   are  not  next  to   their  own
units. However, since each game on  a given map will always start with
the same initial properties (except the starting position of players),
it is  considered that the  map topology  and properties, such  as the
position and quantity of each resource, are known.

\subsection{Terrain Analysis}

A bot  playing StarCraft must  know the  terrain topology and  the map
properties to be able to  do some spacial reasoning. Spacial reasoning
is  necessary to  make  good decisions  such as  where  to expend  for
colonizing an unclaimed part of the  map, where to attack the opponent
in priority, and where defense must be reinforced.

Terrain  analysis  is gathering  and  processing  data about  the  map
topology and properties to  have a qualitative spatial representation,
required to  make spatial  reasoning.  In  StarCraft, map  topology is
characterized among other things by  the terrain height (low, high and
very high, to take \bw terminology,  and a rare fourth level in \lotv)
and  the  nature  of  each   tile  paving  the  map  (walkable  and/or
buildable).  Map properties  include the height and width  of the map,
the  position and  the quantity  of resources,  the possible  starting
position  of  players,  the  presence,  position  and  hit  points  of
destructible obstacles, etc.

One of the main  jobs done by a terrain analysis  is splitting the map
into regions,  and spotting choke points  that are easy to  defend and
make evident region delimiters.

\section{Related work}

There are only a few relevant  papers dealing with terrain analysis in
RTS  games,  and  more  specifically about  StarCraft.   We  can  cite
Perkins'  article in  2010  to  be the  first  significant work  about
terrain   analysis   for    \bw~\cite{Perkins2010},   describing   the
\emph{Brood War Terrain Analysis}  library, or~BWTA.  In 2016, Uriarte
and    Ontañón   have    improved   Perkings'    work   and    propose
BWTA2~\cite{Uriarte2016}.

Some other terrain analysis tools have been developed for \bw, such as
BWEM\footnote{\href{https://github.com/N00byEdge/BWEM-community}{github.com/N00byEdge/BWEM-community}},
and                              its                             clone
\textsc{Overseer}\footnote{\href{https://gitlab.com/OverStarcraft/Overseer}{gitlab.com/OverStarcraft/Overseer}}
for \lotv.  However, none of  them have led  to a publication,  to the
best of  our knowledge,  and they  do not  propose the  computation of
polygons surrounding regions, a useful  feature available in both BWTA
and BWTA2 to get the list of points composing borders of each region.

Since our  work propose  the same  features as  BWTA and  BWTA2, and
share with the latter the same  first steps, we briefly describe these
tools in the following.

In a nutshell, BWTA proposed by Perkins~\cite{Perkins2010} follows
the steps in Alg.~\ref{algo:bwta}:

\begin{algorithm}
    \caption{BWTA}
    \label{algo:bwta}
    \begin{algorithmic}[1] % The number tells where the line numbering should start
      \State Convert the map into a Boolean 2D-vector of walkable tiles.
      \State Detect connected components of obstacles by flood-fill.
      \State Create contour polygons around these connected components.
      \State Compute the Voronoi diagram of walkable lands.
      \State Prune the Voronoi diagram by eliminating  leafs with  a parent
      farther away  from the obstacle  contour than the leaf  itself. This
      minimal distance from  a point to the contour is  called the point's
      \emph{radius}.
      \State Identity nodes   of the Voronoi diagram with  smaller radius than
      their    neighbors.    Those    nodes   are    considered   to    be
      choke point. Regions are then areas on each side of a choke point.
      \State Merge regions  following some  criteria about their areas and the
      radius of their common choke point.  This step is considered by the
      author to  be problematic  since the final  result depends  on which
      order regions have been merged.
      \State Compute choke point polygons.
      \State Compute region polygons.
    \end{algorithmic}
\end{algorithm}

Uriarte and Ontañón's library BWTA2~\cite{Uriarte2016} follows BWTA's main
steps, with some differences in  the algorithm and the implementation.
Algorithmic differences are written in Alg.~\ref{algo:bwta2}.

For detecting connected components of obstacles (Step~2), BWTA2 uses a
contour  tracing algorithm  proposed by  Chang et  al~\cite{Chang2004}
instead of a flood-fill algorithm.  This brings two advantages: 1.  It
allows detecting at the same time the inner and outer contours of each
component.  2.   Each inner tile  is labeled,  allowing us to  know in
constant time to which component a tile belongs to.

Step 4 did  not exist in BWTA. Obstacle contours  are simplified using
the    \texttt{boost::geometry}    library     and    its    algorithm
\texttt{simplify},    which    is    an    implementation    of    the
Ramer-Douglas-Peucker algorithm~\cite{Douglas1973}.

\begin{algorithm}
    \caption{BWTA2}
    \label{algo:bwta2}
    \begin{algorithmic}[1] % The number tells where the line numbering should start
      \State Convert the map.
      \State Detect connected components of obstacles with a contour
        tracing algorithm.
      \State Create contour polygons.
      \State Simplify contours.
      \State Compute the Voronoi diagram.
      \State Prune  the Voronoi  diagram.
      \State Identity nodes.
      \State Merge regions.
      \State Compute choke point polygons.
      \State Compute region polygons.
    \end{algorithmic}
\end{algorithm}

\noindent%
The major differences between BWTA and BWTA2 implementations are:
\begin{itemize}
\item Voronoi diagram are computed using the \texttt{boost} library in
  BWTA2, when BWTA is using  the \texttt{CGAL} library. Compare to the
  latter,  the \texttt{boost}  implementation  build Voronoi  diagrams
  containing more information, like the  coordinates of the centers of
  the  inscribed  circles tangent  to  three  points of  the  obstacle
  contour, which gives us a centroid of each region for free.
\item Pruning the Voronoi diagram  takes advantages of the R-Tree data
  structure~\cite{Guttman1984}    in     BWTA2,    on     which    the
  \texttt{boost::geometry} library  lies on.  Indeed, R-Tree  are data
  structures  used  for  spatial   searching.   Combine  with  packing
  algorithms~\cite{Leutenegger1997,    Garcia1998}   implemented    in
  \texttt{boost::geometry} to  make bulk loading, it  allows answering
  very  quickly  to  many  different queries  about  the  overlapping,
  crossing, etc, of geometrical elements.
\item  Building region  polygons also  takes advantage  of R-Trees  in
  BWTA2,  since it  boils down  to  a query  computing the  difference
  between an obstacle contour with the polygon of each choke point.
\end{itemize}
  
\section{Terrain Analysis as Combinatorial Optimization}

In this paper, we present a new terrain analysis working with both \bw
and \lotv,  named \taunt\footnote{The source code,  experimental setup
  and results can be found at \href{https://github.com/richoux/Taunt/releases/tag/0.1}{github.com/richoux/Taunt/releases/tag/0.1}}.

\subsection{Features}\label{sec:features}

\taunt offers the same features BWTA  and BWTA2 provide: split the map
into regions, mark choke points and base placements, compute contour
polygons of regions and choke points, etc.

The new features  compare to BWTA/BWTA2 are listed below.  To the best
of our knowledge, the two first  ones have not been implemented by any
other terrain analysis tools for StarCraft.
\begin{itemize}
\item Proposing a unique terrain analysis library working with \bw and
  \lotv maps, with the same library interface.
\item Offering different  options of  terrain analysis,  to better  fit
  different needs.
\item Allowing  dynamic terrain analysis, such  as recomputing regions
  after  some  dynamic  event  like the  suppression  of  destructible
  obstacles (also implemented  in by the \bw  terrain analysis library
  BWEM).
\end{itemize}

\taunt stands for ``Terrain Analysis: UNiversal Tool''. Our goal is to
propose a terrain  analysis library which is universal  because it can
be used  for both StarCraft  games, but  also universal because  it is
able to analyze the same map differently, aiming to provide an adapted
analysis to various StarCraft bots.

\subsection{Process}\label{sec:process}

\taunt roughly follows  the same steps as BWTA2 until  its fifth step,
with  a   significant  difference  at   Step~2.   It  also   uses  the
\texttt{boost::geometry}  library,  then  exploiting the  R-Tree  data
structure   for   diverse   queries.    Its  process   is   given   in
Alg.~\ref{algo:taunt}.

\begin{algorithm}
    \caption{\taunt}
    \label{algo:taunt}
    \begin{algorithmic}[1] % The number tells where the line numbering should start
      \State Convert the map.
      \State Detect connected components {\bf of walkable lands} with a contour
        tracing algorithm.
      \State Create contour polygons.
      \State Simplify contours.
      \State Split the  land into areas regarding their  height and if
      they are buildable or not.
      \State Compute contour polygons of unbuildable areas
      (slopes, bridges, etc).
      \State Spot the clusters of resources in each area.
      \State For each area with  more than one resource cluster, run
      the combinatorial optimization problem modeled and solved within
      the framework GHOST\cite{Richoux2016}.
      \State Compute region polygons.
    \end{algorithmic}
\end{algorithm}

One problem  with all existing  terrain analysis libraries for  \bw or
\lotv is that they tend to cut  the map into numerous regions. Some of
these regions  do not  contain any  relevant strategic  properties and
would be  probably not  considered as an  independent region  by human
players.  This  is due  to the  fact that  all existing  libraries are
building regions by  considering each slight narrowing  of the ground,
even large ones, as a choke point.

We  wanted to  counter-balance  this  with \taunt,  and  to propose  a
different way  (actually, different ways) to  cut the map. To  be sure
all regions have some strategical interest, we ensure that all of them
must  contain  exactly one  \emph{resource  cluster},  \ie a  pack  of
mineral  patches,  with  or  without  gas  geyser  around,  with  some
exceptions like islands, areas  surrounded by walkable but unbuildable
lands, and areas surrounded by  areas of different heights but without
any resource clusters (which is very rare in practice).

\begin{figure}[htbp]
	\centering
  \includegraphics[width=0.7\linewidth]{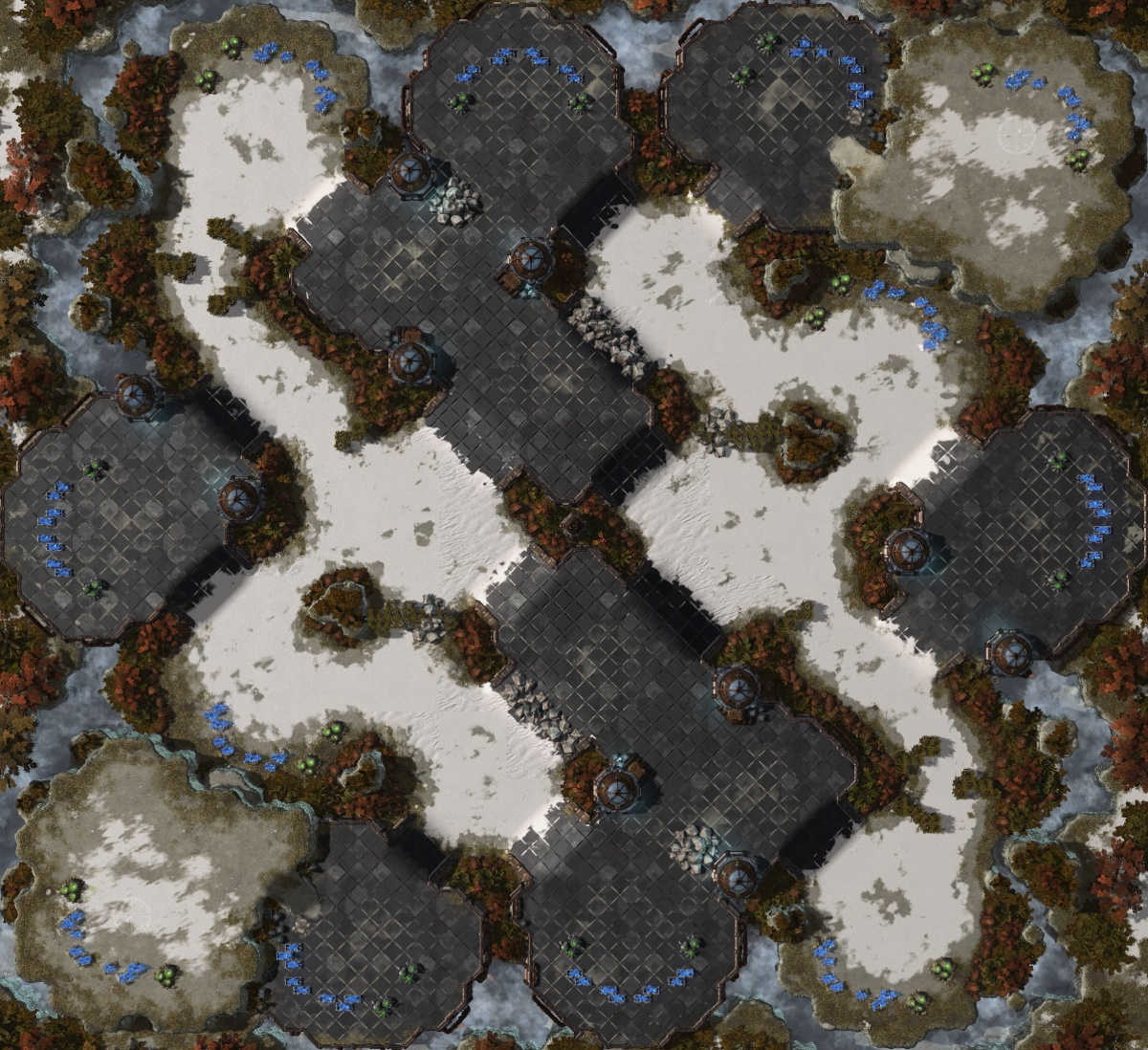}
  \caption{Oxide LE (StarCraft 2 map)}\label{fig:oxide}
\end{figure}

Let's use  the \lotv  map \emph{Oxide  LE} in  Fig.~\ref{fig:oxide} to
illustrate the different steps of Alg.~\ref{algo:taunt}.

At Step~1, \taunt gathers information about the map via the right API,
\ie
BWAPI\footnote{\href{https://github.com/bwapi/bwapi}{github.com/bwapi/bwapi}}
for                   \bw                   maps                   and
CPP-SC2\footnote{\href{https://github.com/cpp-sc2/cpp-sc2}{github.com/cpp-sc2/cpp-sc2}}
for \lotv  maps. Step~2 uses  the same contour tracing  algorithm from
Chang et  al~\cite{Chang2004}, but to compute  connected components of
walkable lands  rather than  obstacles like  in BWTA2.   Although this
algorithm is  from 2004, a  recent survey from He  et al~\cite{He2017}
indicates that  it is  still state-of-the-art for  computing connected
components together with contours.

\begin{figure}[htbp]
	\centering
  \includegraphics[width=0.7\linewidth]{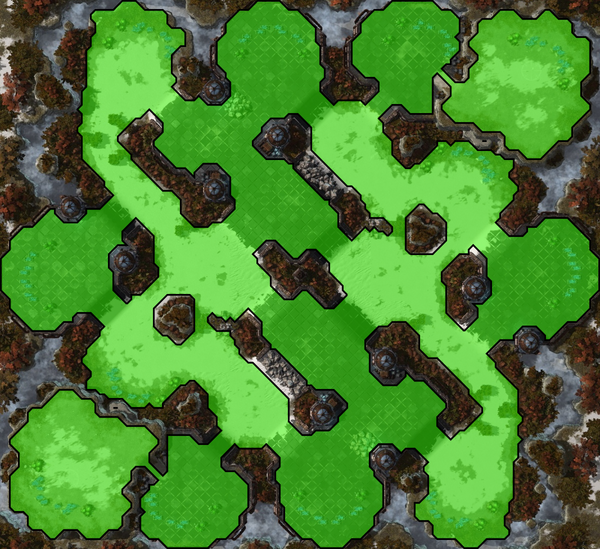}
  \caption{Connected components and contours (Steps 1-4)}\label{fig:oxide_contour} 
\end{figure}

After Step~3 and~4,  we obtain labeled connected  components and their
contour.    The    result   gives    what    we    can   observe    in
Fig.~\ref{fig:oxide_contour},   where   green  parts   are   connected
components (only one in this map) together with their contour in black.

Step~5  consists in  splitting each  component of  walkable land  into
separated areas regarding their height  and if they are unbuildable or
not.   In \lotv,  unbuildable parts  are usually  slopes, bridges  and
woods, and are therefore natural choke points.  In \bw, in addition to
slopes  and bridges,  we can  also  have large  chunks of  unbuildable
areas,  usually located  in  the  central part  of  the  map (see  for
instance  Fig~\ref{fig:circuit_breakers_short}).  Step~6  computes the
contour polygons of these unbuildable areas.

\begin{figure}[htbp]
	\centering
  \includegraphics[width=0.7\linewidth]{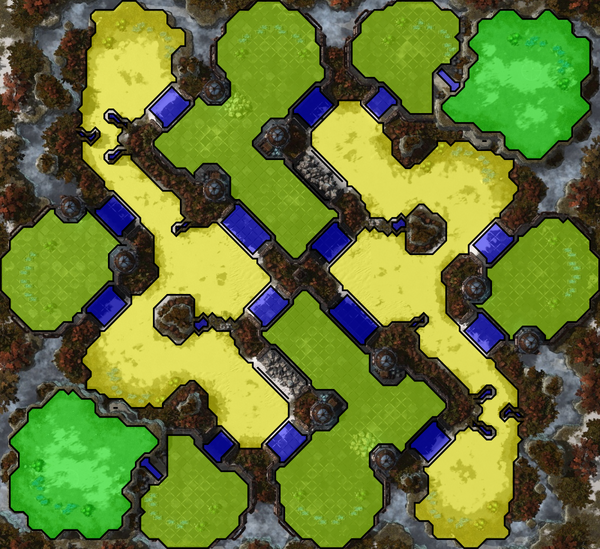}
  \caption{Splitting areas by height (Steps 5-6)}\label{fig:oxide_regions} 
\end{figure}

The     result    of     Steps~5    and~6     can    be     seen    in
Fig.~\ref{fig:oxide_regions}. Blue  parts are unbuildable  areas. Low,
high and very high areas are respectively in yellow, khaki and green.

Since we  aim to have  exactly one  resource cluster per  region (with
some exceptions listed at the beginning  of this section), we need to
locate and  count clusters on each  walkable area. This is  done with
Step~7, and areas with more than  one resource cluster will be handled
by Step~8. Only  buildable areas are concerned by  these steps: since,
by  definition, players  cannot  build bases  upon unbuildable  tiles,
there are never resource clusters on unbuildable areas.

\begin{figure}[htbp]
	\centering
  \includegraphics[width=0.7\linewidth]{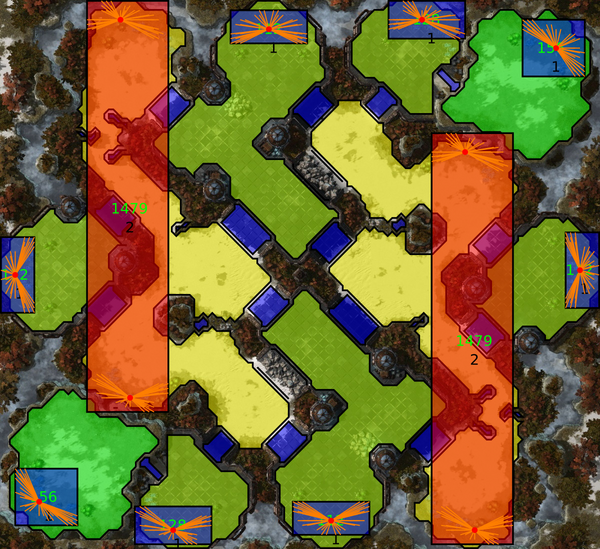}
  \caption{Spotting resource clusters in areas (Step 7)}\label{fig:oxide_resources} 
\end{figure}

In  Fig.~\ref{fig:oxide_resources}, we  can see  resource clusters  in
blue or red boxes. Red dots are  the centroid of a cluster, and orange
lines  connect this  centroid  to  all tiles  upon  which  there is  a
resource of the cluster.

\begin{figure}[htbp]
	\centering
  \includegraphics[width=0.7\linewidth]{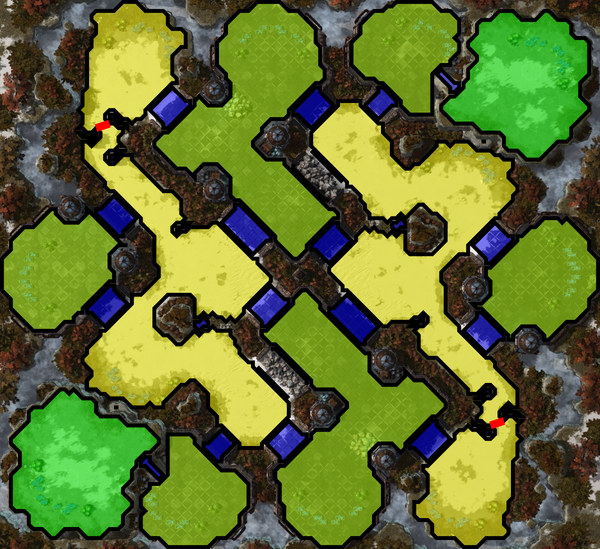}
  \caption{Building region: short separations (Steps 8-9)}\label{fig:oxide_short} 
\end{figure}

\begin{figure}[htbp]
	\centering
  \includegraphics[width=0.7\linewidth]{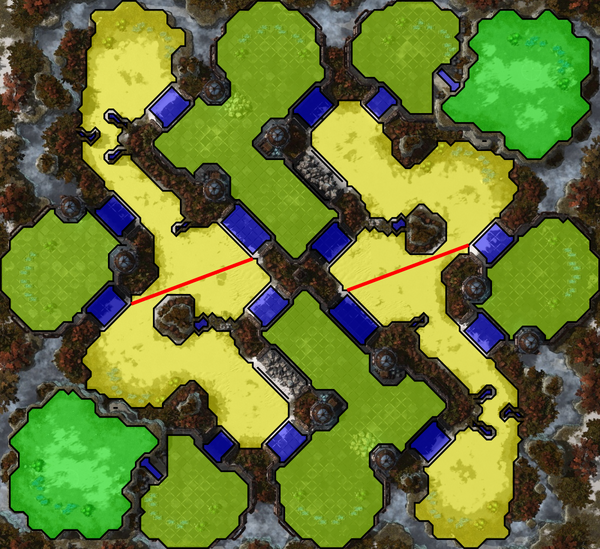}
  \caption{Building region: equivalent surfaces (Steps 8-9)}\label{fig:oxide_areas} 
\end{figure}

If there is only one cluster in an area, it is framed by a blue box on
the figure.   Otherwise, if they  are two  or more clusters,  they are
framed by a red box. In that case, if \(n\) resource clusters exist on
the same  area, we need  to find  \(n-1\) separations to  make regions
containing  exactly  one  cluster  each.   To  do  this,  we  solve  a
combinatorial  optimization  problem   (Step~8)  to  find  separations
satisfying  some   mandatory  constraints  and  optimizing   a  giving
objective  function.   We  show   separations  found  by  solving  two
combinatorial optimization problems while trying to minimize the total
length of  separations (Fig.~\ref{fig:oxide_short}),  corresponding to
choke  points,  and  while  trying   to  minimize  the  least  squares
difference    of    region   surfaces    (Fig.~\ref{fig:oxide_areas}).
Separations are represented by red lines.

Once regions have been computed, it  is trivial to merge two separated
regions by an  obstacle if the latter is destroyed,  and to run Step~8
on this merged region if making  a new separation is required. The way
\taunt decides separations area by area,  allows us to reanalyze a map
locally, without the need to reconsider it as a whole.

\subsection{Intuition behind our combinatorial optimization models}\label{sec:model}

We  modeled the  problem of  placing  \(n-1\) separations  in an  area
containing \(n\) resource clusters to  make \(n\) regions with exactly
one  resource  cluster  each,  with  \(n \geq  2\),  in  a  Constraint
Programming formalism  called Error Function Optimization  Problem, or
EFOP~\cite{Richoux2021}.

An  EFOP model  is  actually an  Error  Function Satisfaction  Problem
model, or EFSP, together with an objective function we aim to maximize
or minimize. An EFSP model is defined  by a set of variables, a set of
domains, \ie the possible values each  variable can take, and a set of
constraints  determining which  combinations  of  variable values  are
possible or forbidden. Our EFSP model is defined as follows:
\cspmodel%
{All possible  separations, \ie  all possible pairs  of points  on the
  contour polygon of an area,}%
{Each  variable   has  the  domain  \(\{0,1\}\),   indicating  if  its
  corresponding separation is selected or not,}%
{\begin{enumerate}\item  \emph{NoCrossings}:  No separations  crossing
    each other, \item  \emph{MaxOneClusterPerRegion}: Separations must
    split  the terrain  into  regions such  that  each region  exactly
    contains one resource cluster.\end{enumerate}}%

In  EFSP  and  EFOP  models,  constraints  are  represented  as  error
functions. For the two constraints in our model, their error function
is expressed as follows:
\begin{itemize}
\item \emph{NoCrossings}: Number of separations crossing each other.
\item  \emph{MaxOneClusterPerRegion}: The  highest number  of resource
  clusters on the  same region minus 1, added to  the number of regions
  without any clusters.
\end{itemize}

Having  models as  small as  possible, \ie  with few  variables, small
domains and few constraints, is important to solve them faster.  We
have the following tricks to reduce the model size:
\begin{itemize}
\item    Ill-formed   separations,    \ie   crossing    resources   or
  unwalkable/unbuildable  tiles,  are filtered  out  from  the set  of
  variables before running the constraint solver.
\item Since  we know how  many separations we  must have to  get \(n\)
  regions, we randomly select \(n-1\)  initial separations and ask the
  solver to handle  the problem as a permutation  problem. This allows
  us  not to  consider a  constraint  checking the  correct number  of
  selected separations.
\end{itemize}

In order to  allow better separations, we enrich the  set of variables
by  adding some  points in  the contour  polygon to  break long  lines
(typically, when  the contour contains  an edge of the  map). Although
this increases the model size by  adding variables, it may allow nicer
separations improving the objective function.

After these variable  reduction tricks and this  enrichment, we obtain
models from  some hundreds to  2,000 variables, regarding the  map and
their areas. The mean numbers of variables among areas within the same
maps are given in Table~\ref{tab:maps}.

We use this EFSP model for  the experiments in this paper. When \taunt
will  be  released,  we  plan  to  propose  other  models  with  other
constraints to offer more  diverse analysis possibilities to StarCraft
bot authors. For instance, we could  add a constraint to force regions
to have a  minimal surface, or to get  perfectly symmetric separations
on symmetrical maps.

We  get an  EFOP model  upon this  EFSP model  by adding  an objective
function. We have tested two different ones:
\objective%
{\begin{enumerate}\item  \emph{MinSeparationLength}: Minimize  the sum
    of separation length. \item \emph{LeastSquaresAreas}: Least squares area difference, \ie minimize the difference of region areas.\end{enumerate}}%

The different region-making we got  with these two objective functions
can        be       observed        in       Fig~\ref{fig:oxide_short}
and~\ref{fig:oxide_areas}, respectively.

\subsection{Our formal combinatorial optimization models}

The EFOP model above is informal and is here to give the intuition. We
show below the formal EFOP model:
\cspmodel%
{\(V = \{v_1, v_2, \ldots, v_k\}\), with \(v_i\) a possible separation}%
{\(D_i = \{0,1\}, \forall i \in [1,k]\)}%
{\(C = \{f_{cross}, f_{clust}\}\)}%
\vspace{-0.3cm}
\begin{IEEEeqnarray*}{rCl}
  f_{cross}(\vec v) & = & \#\{(v_i, v_j) \mid
  cross(v_i, v_j), \forall 1 \leq i,j \leq k, i \neq j\}\\
  f_{clust}(\vec v) & = & Max( 0,
  Max(\#cluster_1,  \ldots, \#cluster_n) - 1 )\\ & &+ \#\{region_j \mid \#cluster_i = 0\}
\end{IEEEeqnarray*}
\noindent%
with  \(\vec  v  =  (v_1,   v_2,  \ldots,  v_k)\),  \(f_{cross}\)  and
\(f_{clust}\) the error  functions representing constraints NoCrossing
and  MaxOneClusterPerRegion,  respectively,  \(cross(v_i,  v_j)\)  the
predicate returning  true iff  \(v_i\) and  \(v_j\) are  crossing each
other,  and \(\#cluster_i\)  the number  of resource  clusters in  the
\(i\)-th region.

Thus, our objectives functions are:
\objective%
{\(O = \{f_{sep}, f_{areas}\}\)}%
\vspace{-0.3cm}
\begin{IEEEeqnarray*}{rCl}
  f_{sep}(\vec v)  & = &  Min \sum_{i=1}^k \left( v_i  \times length_i
  \right)\\
  f_{areas}(\vec v) & = & Min \sum_{i=1}^n\left( mean\_area - area_i \right)^2
\end{IEEEeqnarray*}
\noindent%
with  \(length_i\)  the length  of  the  \(i\)-th separation  \(v_i\),
\(area_i\) the surface  of the \(i\)-th region  and \(mean\_area\) the
mean surface of considered regions.

These   models    have   been   implemented   using    the   framework
GHOST~\cite{Richoux2016}, which contains a local search solver to find
a solution of the problem within  a given timeout. For our experiments
presented   in    Section~\ref{sec:xp},   we   set   a    timeout   of
\(n \times 100\)ms  to find separations in each  area containing \(n\)
resource clusters, with \(n \geq 2\). \taunt contains a mechanism that
double timeouts  and relaunch this  solving step if no  solutions have
been found in its previous run.

\section{Experimental results}\label{sec:xp}

\begin{figure*}[htbp]
	\centering
	\begin{subfigure}[t]{0.24\linewidth}
		\centering
    \includegraphics[width=\linewidth]{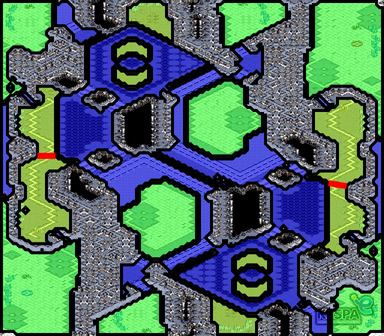}
		\caption{Benzene (StarCraft 1)}\label{fig:benzene_short} 
	\end{subfigure}
  \hfill
	\begin{subfigure}[t]{0.24\linewidth}
		\centering
    \includegraphics[width=\linewidth]{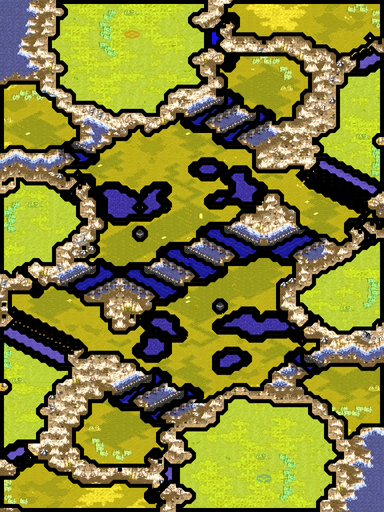}
		\caption{Destination (StarCraft 1)}\label{fig:destination_short} 
	\end{subfigure}
  \hfill
	\begin{subfigure}[t]{0.24\linewidth}
		\centering
    \includegraphics[width=\linewidth]{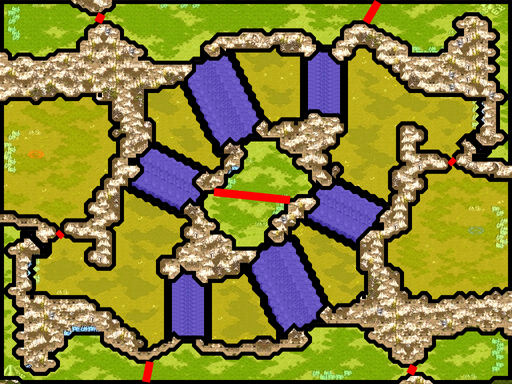}
		\caption{Heartbreak Ridge (StarCraft 1)}\label{fig:heartbreak_ridge_short} 
	\end{subfigure}
  \hfill
	\begin{subfigure}[t]{0.24\linewidth}
		\centering
    \includegraphics[width=\linewidth]{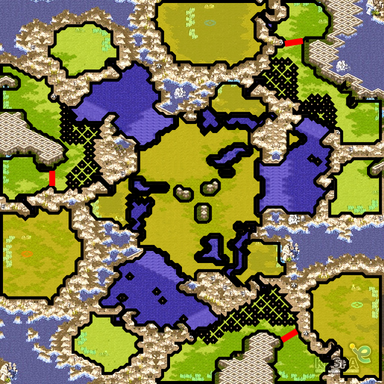}
		\caption{Aztec (StarCraft 1)}\label{fig:aztec_short} 
	\end{subfigure}
  \vskip\baselineskip
	\begin{subfigure}[t]{0.24\linewidth}
		\centering
    \includegraphics[width=\linewidth]{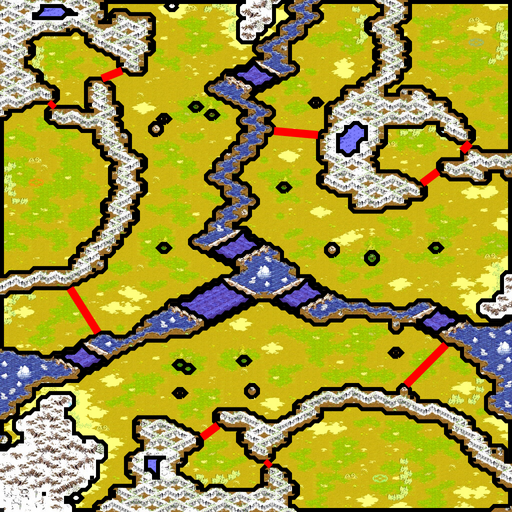}
		\caption{Tau Cross(StarCraft 1)}\label{fig:tau_cross_short} 
	\end{subfigure}
  \hfill
	\begin{subfigure}[t]{0.24\linewidth}
		\centering
    \includegraphics[width=\linewidth]{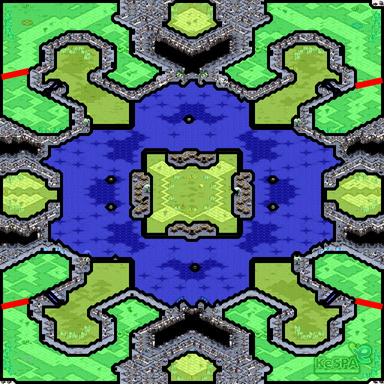}
		\caption{Andromeda (StarCraft 1)}\label{fig:andromeda_short} 
	\end{subfigure}
  \hfill
	\begin{subfigure}[t]{0.24\linewidth}
		\centering
    \includegraphics[width=\linewidth]{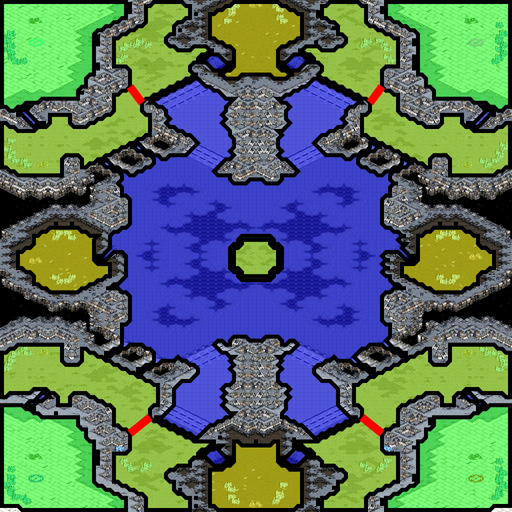}
		\caption{Circuit Breakers (StarCraft 1)}\label{fig:circuit_breakers_short} 
	\end{subfigure}
  \hfill
	\begin{subfigure}[t]{0.24\linewidth}
		\centering
    \includegraphics[width=\linewidth]{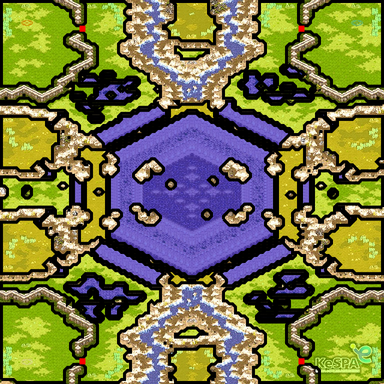}
		\caption{\footnotesize Empire of the Sun (StarCraft 1)}\label{fig:empire_of_the_sun_short} 
	\end{subfigure}
  \vskip\baselineskip
	\begin{subfigure}[t]{0.24\linewidth}
		\centering
    \includegraphics[width=\linewidth]{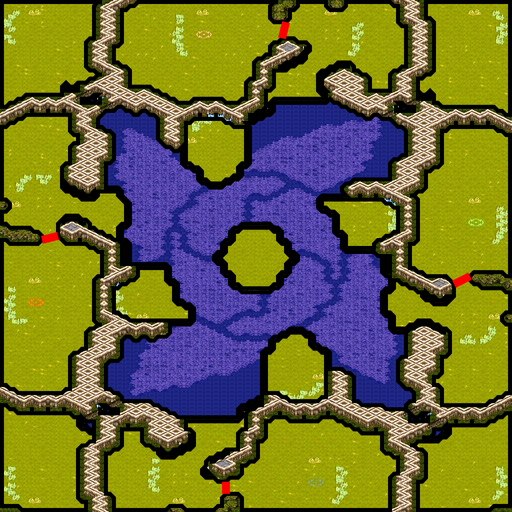}
		\caption{Fortress (StarCraft 1)}\label{fig:fortress_short} 
	\end{subfigure}
  \hfill
	\begin{subfigure}[t]{0.24\linewidth}
		\centering
    \includegraphics[width=\linewidth]{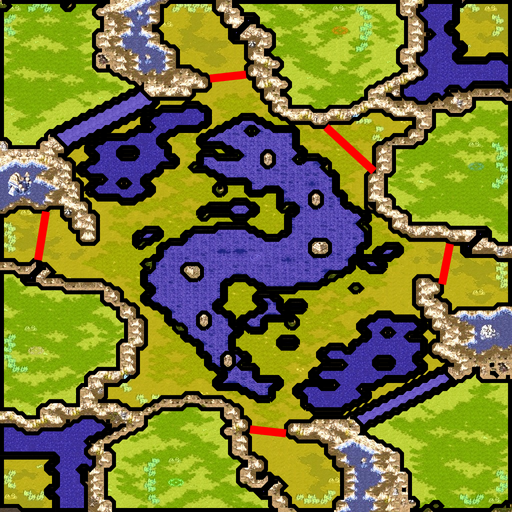}
		\caption{Python (StarCraft 1)}\label{fig:python_short} 
	\end{subfigure}
  \hfill
	\begin{subfigure}[t]{0.24\linewidth}
		\centering
    \includegraphics[width=\linewidth]{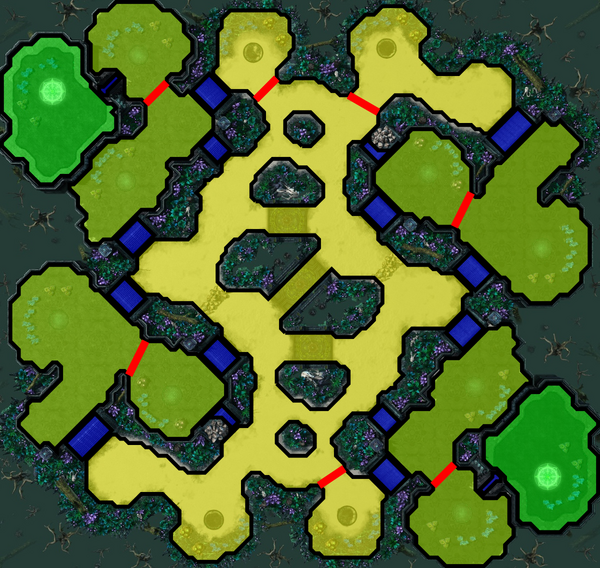}
		\caption{Deathaura LE (StarCraft 2)}\label{fig:deathaura_short} 
	\end{subfigure}
  \hfill
	\begin{subfigure}[t]{0.24\linewidth}
		\centering
    \includegraphics[width=0.93\linewidth]{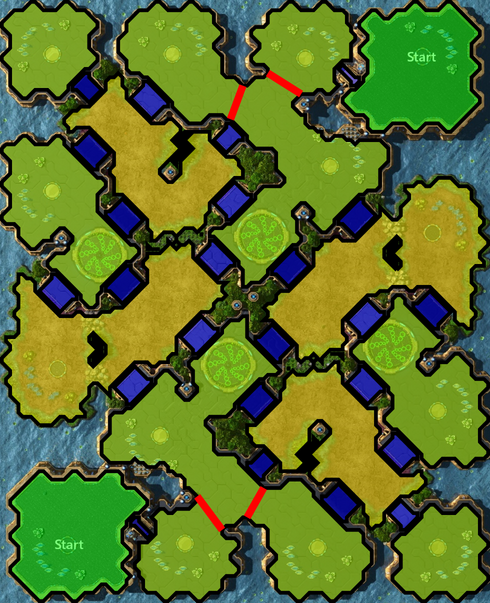}
		\caption{Jagannatha LE (StarCraft 2)}\label{fig:jagannatha_short} 
	\end{subfigure}
  \vskip\baselineskip
	\begin{subfigure}[t]{0.24\linewidth}
		\centering
    \includegraphics[width=\linewidth]{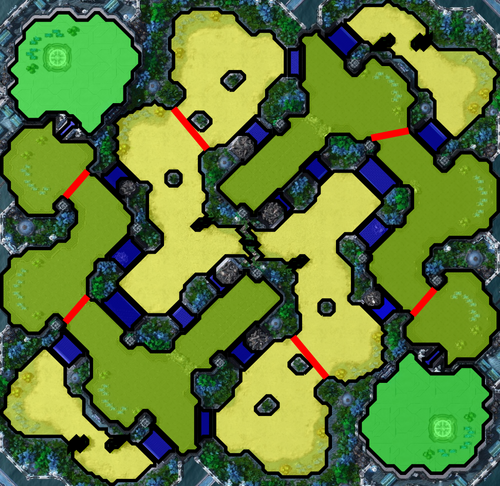}
		\caption{Lightshade LE (StarCraft 2)}\label{fig:lightshade_short} 
	\end{subfigure}
  \hfill
	\begin{subfigure}[t]{0.24\linewidth}
		\centering
    \includegraphics[width=0.93\linewidth]{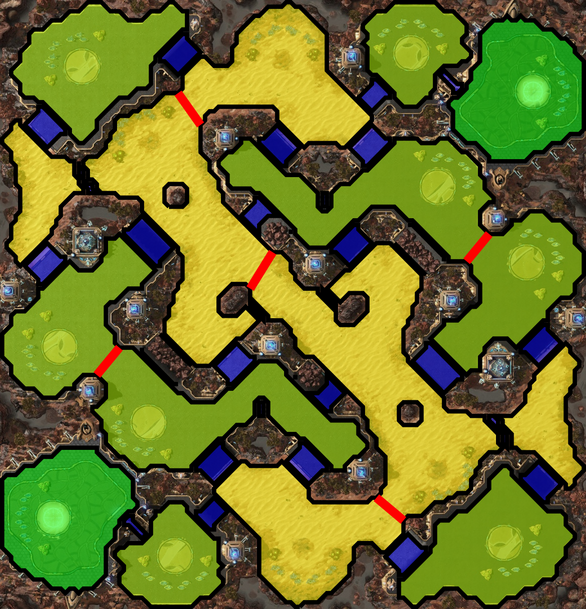}
		\caption{\footnotesize Pillar Of Gold LE (StarCraft 2)}\label{fig:pillar_of_gold_short} 
	\end{subfigure}
  \hfill
	\begin{subfigure}[t]{0.24\linewidth}
		\centering
    \includegraphics[width=\linewidth]{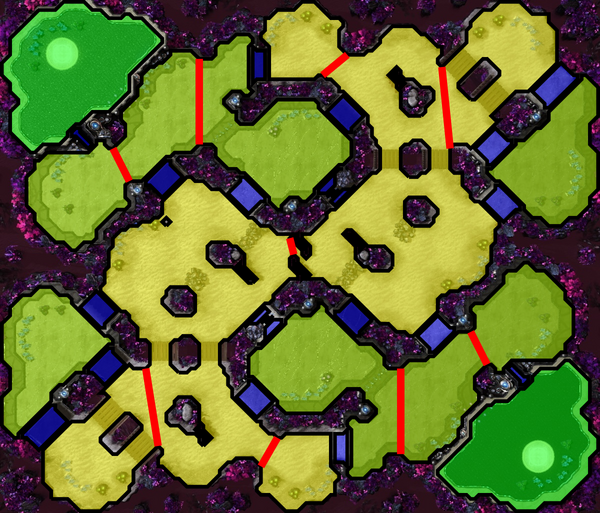}
		\caption{Romanticide LE (StarCraft 2)}\label{fig:romanticide_short} 
	\end{subfigure}
  \hfill
	\begin{subfigure}[t]{0.24\linewidth}
		\centering
    \includegraphics[width=\linewidth]{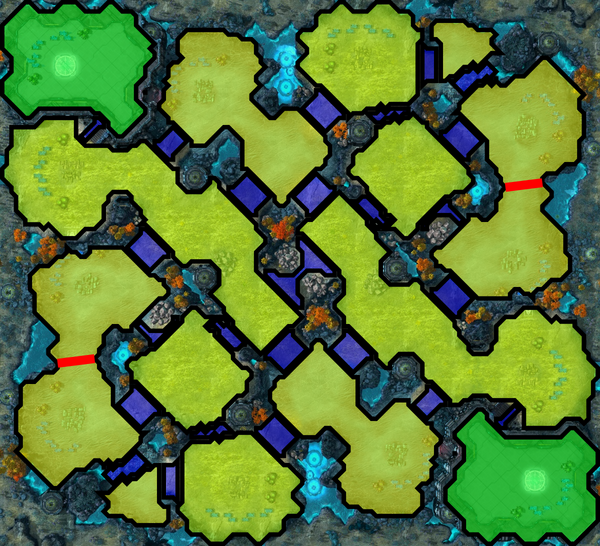}
		\caption{Submarine LE (StarCraft 2)}\label{fig:submarine_short} 
	\end{subfigure}
  \caption{StarCraft maps}
  \label{fig:maps}
\end{figure*}

We  tested our  library over  the 10  \bw maps  from the  annual AIIDE
StarCraft                                                           AI
competition\footnote{\href{https://www.cs.mun.ca/~dchurchill/starcraftaicomp/}{www.cs.mun.ca/~dchurchill/starcraftaicomp}}
and  the 7  \lotv  maps from  the  pre-season 11  of  the SC2AI  Arena
competition\footnote{\href{https://aiarena.net/}{aiarena.net}}.    Map
names as well as the mean number of variables of their EFOP models are
listed  in Table~\ref{tab:maps},  and visual  results considering  the
\emph{MinSeparationLength}   objective    function   are    given   in
Fig.~\ref{fig:maps}, except Oxide  LE since it is  our running example
in Section~\ref{sec:process}.

Tables~\ref{tab:xp}  and~\ref{tab:sc2_runtimes}  give \taunt  runtimes
considering  the   EFOP  model  with   the  \emph{MinSeparationLength}
objective   function,  but   the  \emph{LeastSquaresAreas}   objective
function  leads to  equivalent runtimes.  Unless the  chosen objective
function is computation-heavy, \taunt  runtimes generally depend on the
satisfaction part of the combinatorial problem model.

\begin{table}[htbp]
  \caption{StarCraft maps  used for experiments and  mean number
    of variables of their EFOP models.}
  \begin{center}
    \begin{tabular}{|c|r|c|r|}
      \hline
      \multicolumn{2}{|c|}{\bw maps}&\multicolumn{2}{c|}{\lotv maps}\\
      \hline
      Name & Mean \#vars & Name & Mean \#vars\\
      \hline
      Benzene & 308 & Deathaura LE & 732 \\
      \rowcolor{Gray}
      Destination & 0 & Jagannatha LE & 1434 \\
      Heartbreak Ridge & 673& Lightshade LE & 1103\\
      \rowcolor{Gray}
      Aztec & 292 & Oxide LE & 1285\\
      Tau Cross & 2052& Pillar Of Gold LE & 1001\\
      \rowcolor{Gray}
      Andromeda & 164& Romanticide LE & 1188\\
      Circuit Breakers & 339 & Submarine LE & 454\\
      \hhline{|~~|--}
      \rowcolor{Gray}
      Empire of the Sun & 508 & \multicolumn{2}{c}{\cellcolor{white}} \\
      Fortress & 495 & \multicolumn{2}{c}{(LE: Ladder Edition)}\\
      \rowcolor{Gray}
      Python & 1063 & \multicolumn{2}{c}{\cellcolor{white}}\\
      \cline{1-2}
    \end{tabular}
    \label{tab:maps}
  \end{center}
\end{table}

\subsection{Comparaison BWTA2}

Tests have been  conducted using a bot under development  able to play
both at \bw  and \lotv, on a  machine equipped with a Core  i9 9900 CPU
and 32 GB of RAM, running on  Windows 10. \taunt is written in C++ and
has been  compiled with the Visual  Studio 2022 compiler with  the /02
optimization option.

The  terrain  analysis  results,  namely  the  region-making  and  the
computation of  their contour  polygon, are good  on all  tested maps:
\taunt shows its ability to split  a map into regions with the desired
properties.  In order  to have  a  point of  comparison with  existing
libraries and  an objective  metric for  the evaluation  of $\taunt$'s
performance, we compare  runtimes of BWTA2 and \taunt on  \bw maps. We
choose BWTA2  rather than a more  modern library like BWEM,  since the
latter   does   not  propose   features   like   the  computation   of
contours. BWTA2 is  actually the more modern  terrain analysis library
for \bw offering equivalent features as \taunt.

\begin{table}[htbp]
  \caption{Runtime comparison in milliseconds between BWTA2 and \taunt.}
  \begin{center}
    \begin{tabular}{|c|r|r|}
      \hline
      \textbf{\bw Map}& \textbf{BWTA2}& \textbf{\taunt} \\
      \hline
      Benzene & 906~~~\, & 344~~~\, \\
      % Destination & 843 & 15~~~\, \\
      \rowcolor{Gray}
      Heartbreak Ridge & 794~~~\, & 821~~~\, \\
      Aztec & 1112~~~\, & 751~~~\, \\
      \rowcolor{Gray}
      Tau Cross & 1046~~~\, & 1839~~~\, \\
      Andromeda & 1114~~~\, & 626~~~\, \\
      \rowcolor{Gray}
      Circuit Breakers & 1187~~~\, & 654~~~\, \\
      Empire of the Sun & 1132~~~\, & 842~~~\, \\
      \rowcolor{Gray}
      Fortress & 1074~~~\, & 491~~~\, \\
      Python & 1089~~~\, & 1203~~~\, \\
      \hline
      Median runtime & 1089~~~\, & 751~~~\, \\
      Mean runtime & 1050.44 & 841.22\\
      Pop. standard deviation & 116.34 & 419.82\\
      \hline
    \end{tabular}
    \label{tab:xp}
  \end{center}
\end{table}

Runtime comparisons are shown in Table~\ref{tab:xp}.  BWTA2 and \taunt
have been run on the same machine through the same bot, under the same
conditions.  The full  run of BWTA2 and \taunt  algorithms is measured
in milliseconds, including fetching data  from APIs and preparing data
structures.  The reader can observe that the map Destination is not on
this table.   We have taken out  this map from comparisons  because it
would  be completely  at  the  advantage of  \taunt,  being an  unfair
comparison with  BWTA2.  Indeed,  Destination is a  map such  that all
walkable areas contain one resource  cluster only, so there is nothing
to compute for \taunt at Step~8 in Alg.~\ref{algo:taunt}, which is the
more  complex  and  computation  heavy  step.  This  is  why  we  have
0~variables  in  Table~\ref{tab:maps},  and   why  there  are  no  red
separation  lines  in  Fig~\ref{fig:destination_short}. As  a  result,
Destination is processed in 15ms by \taunt, against 843ms by BWTA2.

Results from Table~\ref{tab:xp} allow us  to conclude that \taunt runs
faster over most  maps than BWTA2, and has a  significant lower median
and mean  runtime.  The standard  deviation is almost 4  times higher,
though: this is due to  the runtime of Step~8 in Alg.~\ref{algo:taunt}
that  depends on  the number  of  areas to  handle and  the number  of
separations to found.

\begin{table}[htbp]
  \caption{Runtimes in milliseconds on \lotv maps.}
  \begin{center}
    \begin{tabular}{|c|r|}
      \hline
      \textbf{\lotv maps}& \textbf{\taunt} \\
      \hline
      Deathaura LE& 1283~~~\, \\
      \rowcolor{Gray}
      Jagannatha LE& 431~~~\, \\
      Lightshade LE& 755~~~\, \\
      \rowcolor{Gray}
      Oxide LE& 377~~~\, \\
      Pillar Of Gold LE& 824~~~\, \\
      \rowcolor{Gray}
      Romanticide LE& 3475~~~\, \\
      Submarine LE& 242~~~\, \\
      \hline
      Median runtime & 755~~~\, \\
      Mean runtime & 1055.28\\
      Pop. standard deviation & 1039.35\\
      \hline
    \end{tabular}
    \label{tab:sc2_runtimes}
  \end{center}
\end{table}

Since there are no terrain  analysis library for \lotv with equivalent
features as \taunt, we could not  make any runtime comparison on \lotv
maps.   Thus, Table~\ref{tab:sc2_runtimes}  shows runtimes  for \taunt
only. Like for Table~\ref{tab:xp}, runtimes take into account the full
run   of  $\taunt$'s   algorithm,   including   building  inputs   and
initializing    data    structures.     Looking    at    medians    in
Tables~\ref{tab:xp} and~\ref{tab:sc2_runtimes},  we can  conclude that
runtimes  over  \bw and  \lotv  maps  are  equivalent. The  \lotv  map
Romanticide is  an exception: it  contains 2 areas with  2 separations
required  for each,  and  1  area requiring  5  separations, which  is
significantly more than  the average map.  This is the  only map where
\taunt needed to use its inner  mechanism that doubles the timeout and
relaunch the solver,  since no satisfying separations  have been found
during the first  run of the solver.  This explains  the difference of
runtimes between Romanticide and other \lotv maps.

\section{Conclusion and perspectives}

In this paper, we propose a library, \taunt, handling terrain analysis
in StarCraft games  as a combinatorial optimization  problem.  This is
the first terrain analysis work for StarCraft to tackle the problem as
a combinatorial optimization problem, and  the first to use Constraint
Programming to model  and solve it.  We give the  problem model in the
Error Function Optimization Problem formalism, and we test our library
on 10 \bw maps and 7 \lotv maps from StarCraft AI competitions.

The main features of our library  are: 1.  The possibility for \bw and
\lotv bots  to use it  with the same interface,  2. The split  of maps
into  regions in  a  human-like fashion,  avoiding  creating too  many
regions without  significant strategical values, 3.   The computing of
region contours, 4.  The ability  to dynamically reanalyze the terrain
(for  instance  after  the  destruction   of  an  obstacle),  5.   The
possibility to  make different  kind of terrain  analysis of  the same
map. At the moment this article  is written, no other terrain analysis
libraries for StarCraft games propose all these features.

Concerning this last feature, we show  in this paper two ways to split
a  map into  regions, according  to the  model objective  function.  A
future work  to enhance the  library can consist in  implementing more
diverse  combinatorial optimization  problem models  to fit  different
needs  in terms  of terrain  analysis.   For instance,  we could  have
models with a constraint to give  a maximal or minimal bound of region
surfaces,  to  force  symmetrical  regions   (if  the  map  is  itself
symmetrical), or to allow large, central buildable regions without any
resource clusters.

We strongly  believe that having  a universal tool to  get qualitative
spatial  representations,   able  to   handle  both   StarCraft  games
homogeneously  and  offering  the   possibility  of  making  different
analyses of  the same map, is  a necessary step toward  bots with true
adaptability,  able to  make non-scripted  decisions according  to the
environment and the current situation in a  game. We hope to see a new
wave of bots emerges thanks to \taunt.

\bibliographystyle{IEEEtran}
\bibliography{IEEEfull,taunt}

\end{document}